\documentclass[sigconf]{acmart}
\copyrightyear{2026}
\acmYear{2026}
\setcopyright{cc}
\setcctype{by}
\acmConference[MM '26]{Proceedings of the 34th ACM International Conference on Multimedia}{November 10--14, 2026}{Rio de Janeiro, Brazil}
\acmBooktitle{Proceedings of the 34th ACM International Conference on Multimedia (MM '26), November 10--14, 2026, Rio de Janeiro, Brazil}
\acmDOI{10.1145/3767308.3835167}
\acmISBN{979-8-4007-2213-4/2026/11}

\acmSubmissionID{1510}

\AtBeginDocument{%
  }

\usepackage{amsmath}
\usepackage{colortbl}
\usepackage{subcaption}
\usepackage{caption}
\usepackage[capitalize,nameinlink]{cleveref}

\begin{document}

\title[CurvSpec: Adaptive Multi-Curvature Learning for Partial Relevant Video Retrieval]{CurvSpec: Adaptive Multi-Curvature Learning \\ for Partial Relevant Video Retrieval}

\author{Zhen Liu}
\authornote{Contributed equally to this research.}
\affiliation{%
  \institution{Tongji University}
  \city{Shanghai}
  \country{China}
}
\email{sincetodayz@gmail.com}

\author{Letian Li}
\authornotemark[1]
\affiliation{%
  \institution{SIGS, Tsinghua University}
  \city{Shenzhen}
  \country{China}
}
\email{lilt24@mails.tsinghua.edu.cn}

\author{Jinpeng Wang}
\affiliation{%
  \institution{Harbin Institute of Technology, Shenzhen}
  \city{Shenzhen}
  \country{China}
}
\email{wangjp26@gmail.com}

\author{Shuzhao Xie}
\affiliation{%
  \institution{SIGS, Tsinghua University}
  \city{Shenzhen}
  \country{China}
}
\email{xsz24@mails.tsinghua.edu.cn}

\author{Yuzhi Huang}
\affiliation{%
  \institution{SIGS, Tsinghua University}
  \city{Shenzhen}
  \country{China}
}
\email{yz_huang13@163.com}

\author{Jingyan Jiang}
\authornote{Corresponding authors.}
\affiliation{%
  \institution{SIGS, Tsinghua University}
  \city{Shenzhen}
  \country{China}
}
\email{jiangjingyanjlu@gmail.com}

\author{Zhi Wang}
\authornotemark[2]
\affiliation{%
  \institution{SIGS, Tsinghua University}
  \city{Shenzhen}
  \country{China}
}
\email{wangzhi@sz.tsinghua.edu.cn}

\renewcommand{\shortauthors}{Zhen Liu et al.}


\begin{abstract}
   Partially Relevant Video Retrieval (PRVR) seeks to retrieve untrim-med videos containing a moment that matches a text query, without temporal annotations.
   The relevant moment may last only seconds within a video spanning several minutes, creating an extremely low signal-to-noise ratio that makes PRVR more challenging than standard full-video retrieval.
   This task presents two intertwined challenges:
   (1) \emph{signal dilution}, where coarse global representations blur the brief relevant signal into the dominant irrelevant surroundings;
   (2) \emph{curvature rigidity}, where embedding all videos in the same fixed-geometry space distorts representations for videos that range from flat atomic events to deep compositional hierarchies.
   Existing PRVR methods have improved moment selection and cross-modal matching, but they still typically encode all videos in a single fixed-curvature retrieval space, limiting their ability to model diverse video structures.
   To address both challenges, we propose Curv\-Spec, a framework that learns content-adaptive curvature for video retrieval representations rather than imposing a fixed geometric prior.
   Curv\-Spec processes features through parallel Euclidean and hyperbolic attention layers, with independently learned curvatures assigned to the hyperbolic layers, and a content-aware fusion mechanism routes each input to its most suitable geometric regime.
   To further suppress signal dilution, Curv\-Spec represents each video with semantic centroids whose number is determined by the video's content complexity, projects them onto the learned manifold, and matches each query against its nearest centroid by geodesic distance.
   Experiments on ActivityNet Captions, TVR, and Charades-STA demonstrate state-of-the-art retrieval performance.
\end{abstract}




\maketitle

\section{Introduction}
\label{sec:intro}

The rapid growth of online video has driven extensive research on text-to-video retrieval~\cite{ibrahimi2023audio, tian2024holistic, RIVRL, wu2023cap4video, luo2022clip4clip, wu2024cap4video_plus, wang2023align_tell, li2026revisiting, li2026imagine}, a task that seeks to bridge the semantic gap between natural-language descriptions and visual content.
Standard methods assume that a query describes the entire video, yet real-world untrimmed videos often contain content only partially relevant to a given query.
\textit{Partially Relevant Video Retrieval}~(PRVR)~\cite{dong2022prvr} formalizes this more realistic setting: given a text query, retrieve untrimmed videos that contain at least one semantically matching moment.
The relevant moment may last only seconds within a video spanning several minutes, creating an extremely low signal-to-noise ratio that distinguishes PRVR from standard full-video retrieval.

\textbf{\begin{figure*}[t]
  \centering
  \includegraphics[width=\linewidth]{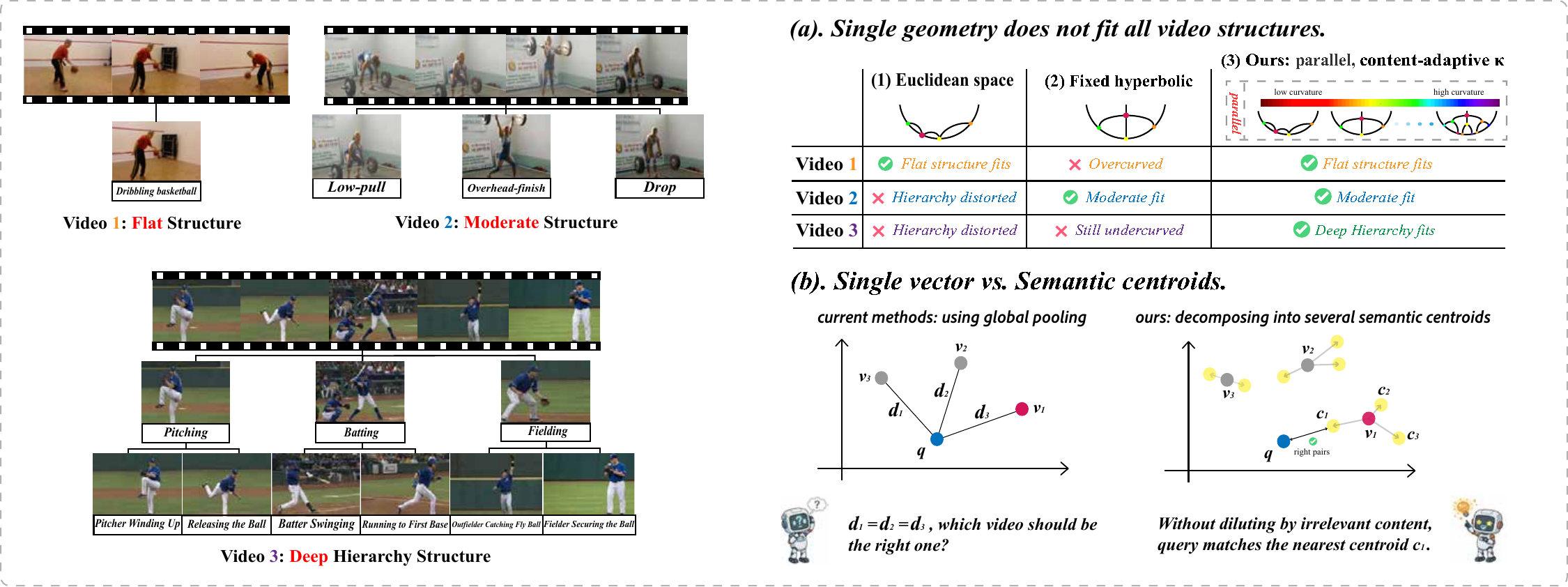}
  \caption{\textbf{Motivation and key ideas.}
  (a)~Videos exhibit diverse structures (Video 1 to 3, from flat to deep hierarchies). Neither Euclidean space nor a single fixed hyperbolic curvature accommodates all of them; CurvSpec routes each video to its optimal geometry.
  (b)~Global pooling dilutes relevant signals into an ambiguous single vector; semantic centroids decompose the video into coherent regions, allowing the query to match the nearest relevant centroid free from irrelevant interference.}
  \label{fig:teaser}
\end{figure*}}

An important challenge in PRVR is \textbf{signal dilution}: coarse global representations blur the brief relevant signal into the dominant irrelevant surroundings, making fine-grained, segment-aware matching essential.
Existing methods~\cite{dong2022prvr, wang2023gmmformer, wang2024gmmformerv2uncertaintyawareframework} have made progress through improved moment selection and cross-modal matching strategies, yet they share a common assumption: the embedding space is geometrically adequate for all videos.
The quality of any matching function, however, is bounded by the fidelity of its input representations, which depends not only on the encoder architecture but also on the geometric properties of the embedding space.

Among these geometric properties, we identify \textbf{curvature rigidity}, the practice of forcing all videos into the same fixed-geometry space regardless of their structural depth, as an underappreciated source of degradation.
Real-world videos range from flat atomic events to deep multi-level hierarchies~(\cref{fig:teaser}a), and these structures have intrinsically different geometries: flat relationships are naturally accommodated in Euclidean space, while deep hierarchies are more faithfully captured by hyperbolic manifolds, whose exponential volume growth mirrors the branching structure of trees~\cite{nickel2017poincare}.
Hyperbolic geometry provides the volume needed for hierarchical videos but imposes spurious separation on flat ones; Euclidean geometry preserves uniform distances but compresses hierarchical depth.
No single curvature resolves this tension, committing to one can distort the representations of videos it does not suit.

In PRVR, curvature rigidity can further exacerbate signal dilution.
If the embedding geometry distorts distances around the relevant moment, that moment may rank lower than irrelevant segments that happen to sit in a geometrically favorable region~(\cref{fig:teaser}b).
Recent hyperbolic methods~\cite{wen2025hover, Li25_HLFormer} show the value of non-Euclidean geometry for video retrieval, yet impose a single fixed curvature on all inputs, biasing every video toward the same geometric regime regardless of actual structural depth.
Meanwhile, most methods still compress each video into a single global vector, conflating relevant moments with surrounding irrelevant content and leaving retrieval further exposed to segments sharing only surface-level similarity with the query.

These observations call for a framework that \emph{(i)}~adapts its embedding geometry to each video's structural depth, and \emph{(ii)}~explicitly isolates relevant content from irrelevant interference.
We address both in \textbf{CurvSpec}, built on two designs.
(1)~To overcome \textbf{curvature rigidity}~(\cref{fig:teaser}a), we introduce a \emph{learnable curvature spectrum}: $N$ Euclidean and $N$ hyperbolic attention layers operating in parallel, with each hyperbolic layer using an independently learned curvature.
A content-aware fusion mechanism weights each layer's contribution according to the input, routing structurally simple videos toward Euclidean layers and hierarchically complex ones toward hyperbolic layers.
The hyperbolic curvatures are optimized end-to-end, eliminating the need for manual geometric tuning.
(2)~To combat \textbf{signal dilution}~(\cref{fig:teaser}b), we decompose each video into multiple \emph{semantic centroids}, each capturing a coherent thematic region.
The centroids are refined through cross-attention over frame-level and clip-level temporal representations, so each centroid reflects a semantically distinct segment rather than a blend of unrelated content.
By projecting the refined centroids onto the learned hyperbolic manifold, retrieval reduces to matching a query against the nearest centroid, naturally suppressing interference from irrelevant content.

Extensive experiments on ActivityNet Captions, TVR, and Char\-ades-STA demonstrate that CurvSpec achieves competitive performance at computational cost comparable to single-geometry baselines.
Further analysis confirms that the learned curvatures diversify from identical initialization and allocate geometric capacity in proportion to structural complexity, with retrieval gains disproportionately concentrated on structurally complex videos.

Our contributions are as follows:
\begin{itemize}
    \item We identify curvature rigidity as a distinct bottleneck in PRVR and address it with a learnable curvature spectrum of parallel Euclidean and hyperbolic attention layers, where each hyperbolic layer uses an independently optimized curvature and a content-aware mechanism routes each video to its most suitable geometric regime.
    \item We address signal dilution through semantic centroid decomposition, which represents video as a set of coherent thematic regions refined through cross-attention, reducing partial relevance matching to nearest-centroid retrieval that isolates relevant moments from irrelevant content.
    \item Extensive experiments on three benchmarks establish state-of-the-art retrieval performance, and analysis confirms that the learned curvatures allocate geometric capacity in proportion to structural complexity, with gains disproportionately concentrated on structurally complex videos.
\end{itemize}

\section{Related Work}

\subsection{Partially Relevant Video Retrieval}
Text-to-video retrieval (T2VR)~\cite{Chen2020FineGrainedVR, gabeur2020mmt, luo2022clip4clip, wu2024cap4video_plus, wang2023align_tell, zhao2022centerclip} typically assumes global video--query alignment, which is inadequate for untrim-med videos containing only partially relevant content.
Dong et al.~\cite{dong2022prvr} formalized this setting as Partially Relevant Video Retrieval (PRVR), where a query may match only a temporal segment rather than the entire video.
Early PRVR methods alleviate this mismatch by generating multi-scale segment representations~\cite{dong2022prvr} or modeling temporal context with Gaussian-weighted attention~\cite{wang2023gmmformer, wang2024gmmformerv2uncertaintyawareframework}.
Recent work further improves partial matching through CLIP-based distillation~\cite{Dong2025DualLW}, event-level alignment~\cite{Jiang2023ProgressiveEA}, uncertainty modeling~\cite{zhang2025enhancingpartiallyrelevantvideo}, ambiguity-aware contrastive learning~\cite{cho2025arl}, and active or uneven event modeling~\cite{song2025efficientpartiallyrelevantvideo, zhu2025uneveneventmodelingpartially}.
These methods have substantially advanced PRVR by locating or emphasizing more informative temporal regions.
However, most still rely on global or fixed-geometry representations, which can dilute short relevant moments and fail to accommodate heterogeneous video structures.
CurvSpec addresses these limitations by combining adaptive retrieval geometry with semantic centroid decomposition.

\subsection{Hyperbolic Representation Learning}
Hyperbolic spaces are well suited to hierarchical and tree-structured data because their exponential volume growth naturally preserves branching relations~\cite{nickel2017poincare, Khrulkov_2020_CVPR, hyper_survey1, hyper_survey2, peng2022hyperbolic_survey}.
Several formulations have been developed for stable and discriminative hyperbolic learning, including Lorentzian embeddings and distance learning~\cite{nickel2018learningcontinuoushierarchieslorentz, law2019lorentzian}.
These ideas have also been extended to multimodal representation learning, such as image--text hierarchy modeling~\cite{desai2023meru}, hierarchical video understanding~\cite{wen2025hover}, and compositional CLIP representations~\cite{PalSDFGM2024}.
For video retrieval, hyperbolic geometry is attractive because videos often contain nested event structures, from atomic actions to multi-step activities.
In PRVR, HLFormer~\cite{Li25_HLFormer} introduces hyperbolic geometry to encode such hierarchical video structures, but uses a single fixed curvature for all inputs.
This fixed choice cannot adapt to videos with different structural depths.
Our work instead learns a curvature spectrum and adaptively fuses multiple geometric regimes according to video content.

\begin{figure*}[t]
    \centering
    \includegraphics[width=0.95\linewidth]{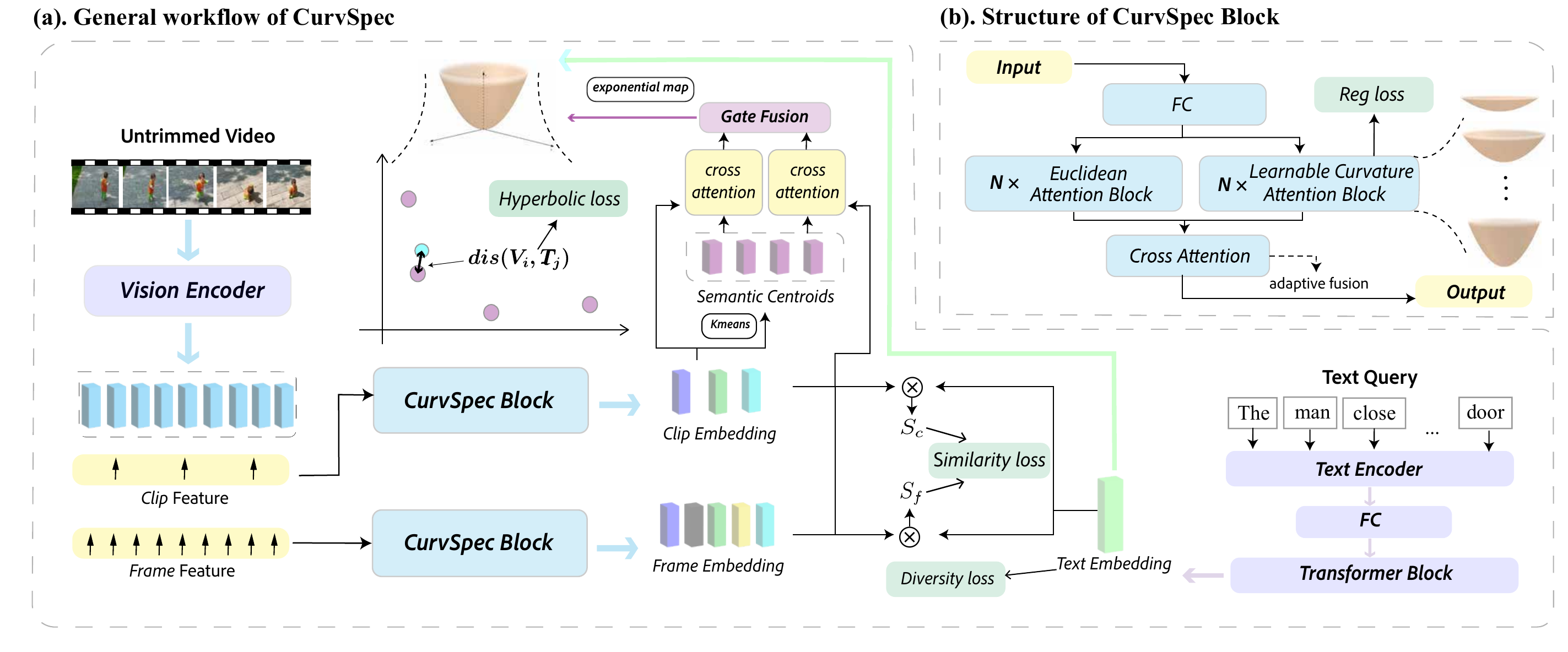}
    \caption{\textbf{Overview of CurvSpec.}
    (a)~Text queries and video frames/clips are encoded and processed by the CurvSpec block to yield enriched representations $\mathbf{Z}_c$ and $\mathbf{Z}_f$.
    Clip features are further decomposed into semantic centroids, which, together with the query, are projected onto the Lorentz manifold for fine-grained partial relevance matching.
    (b)~The CurvSpec block processes input through parallel Euclidean and Lorentz attention branches, each at a distinct geometry.
    A content-aware fusion mechanism adaptively combines the multi-geometry representations.}
    \label{fig:methods}
\end{figure*}

\section{Method}
\label{sec:method}

\subsection{Preliminaries}
\label{sec:background}

\noindent\textbf{Problem formulation.}\quad
Given a dataset $\mathcal{D} = \{(V_i, T_i)\}$, each untrimmed video $V_i$ is associated with text descriptions  
$T_i = \{t_i^1, \ldots, \\ t_i^{m_i}\}$, where each $t_i^j$ corresponds to a specific temporal moment within $V_i$.
Temporal boundaries are not annotated.
The goal of PRVR is to retrieve, for a given text query, videos containing a semantically matching moment, despite the query being relevant to only a fraction of the video content.

\noindent\textbf{The Lorentz model.}\quad
Our framework operates in a mixed geometry that combines Euclidean and hyperbolic spaces.
We adopt the Lorentz model, in which the hyperboloid $\mathbb{H}_{\kappa}^{d} = \{x \in \mathbb{R}^{d+1} : \langle x, x \rangle_{\mathcal{L}} = -\tfrac{1}{\kappa},\; x_0 > 0\}$ is equipped with the Minkowski inner product:
\begin{equation}
    \langle x, y \rangle_{\mathcal{L}} \;=\; -x_0 y_0 + \textstyle\sum_{i=1}^{d} x_i y_i.
    \label{eq:minkowski}
\end{equation}
The geodesic distance between two points on the manifold is:
\begin{equation}
    d_\kappa(x, y) \;=\; \tfrac{1}{\sqrt{\kappa}}\;\operatorname{arcosh\ }\bigl({-}\kappa\,\langle x, y \rangle_{\mathcal{L}}\bigr).
    \label{eq:geodesic}
\end{equation}
Transitioning between the tangent space $T_{o}\mathbb{H}_\kappa^d$ at the origin $o = \bigl(\tfrac{1}{\sqrt{\kappa}},\,0\bigr)$ and the manifold is achieved via the exponential and logarithmic maps.
The exponential map traces the geodesic from $o$ in the direction of a tangent vector $v \in T_{o}\mathbb{H}_\kappa^d$:
\begin{equation}
    \exp_{o}^{\kappa}(v) \;=\; \cosh\Bigl(\sqrt{\kappa}\,\|v\|_2\Bigr)\,o \;+\; \tfrac{1}{\sqrt{\kappa}}\;\sinh\Bigl(\sqrt{\kappa}\,\|v\|_2\Bigr)\,\tfrac{v}{\|v\|_2},
    \label{eq:expmap}
\end{equation}
and its inverse recovers the tangent representation of a manifold point $x \in \mathbb{H}_\kappa^d$:
\begin{equation}
\begin{aligned}
    \log_{o}^{\kappa}(x) &\;=\; d_\kappa(o,\,x)\;\frac{x_{\perp}}{\|x_{\perp}\|_2}, \\
    x_{\perp} &\;=\; x + \kappa\,\langle o,\, x\rangle_{\mathcal{L}}\,o,
\end{aligned}
    \label{eq:logmap}
\end{equation}
where $x_{\perp}$ denotes the component of $x$ orthogonal to $o$ under the Minkowski metric.
Curvature $\kappa$ governs the rate of exponential volume growth: larger $\kappa$ allocates more representational capacity to deep hierarchies, while $\kappa \!\to\! 0$ recovers flat Euclidean geometry.
This geometric sensitivity motivates learning a distinct $\kappa$ per attention layer, allowing the model to discover a curvature spectrum that reflects the varied geometric demands of the data rather than relying on a single prescribed geometry.

\subsection{Overall Architecture}
\label{sec:framework}

As illustrated in~\cref{fig:methods}a, CurvSpec follows a dual-stream encoding paradigm~\cite{song2025efficientpartiallyrelevantvideo, Li25_HLFormer, wang2023gmmformer, wang2024gmmformerv2uncertaintyawareframework}.
Text queries and video content are independently encoded, then compared via multi-granularity similarity.

\noindent\textbf{Text encoding.}\quad
A text query $T = \{w_1, \ldots, w_S\}$ is encoded by a frozen RoBERTa~\cite{liu2019robertarobustlyoptimizedbert} to obtain $X_q \in \mathbb{R}^{S \times d_q}$.
These representations are linearly projected, augmented with learnable positional embeddings $P_\text{text}$, and refined by Euclidean self-attention~\cite{vaswani2023attentionneed}:
\begin{equation}
    Z_q = \mathrm{Attn}_\text{E}\bigl(\mathrm{FC}(X_q) + P_\text{text}\bigr),
\end{equation}
followed by attention-weighted pooling into a single query vector:
\begin{equation}
    q = \textstyle\sum_{s=1}^{S} a_s\,z_{q,s},
    \label{eq:query_pool}
\end{equation}
where $a_s = \mathrm{softmax}_s\,(w_q^\top z_{q,s})$, $w_q \in \mathbb{R}^d$ is a learnable projection, and $\{z_{q,s}\}$ are the refined token representations.

\noindent\textbf{Video encoding.}\quad
An untrimmed video is represented at two temporal granularities: $L$ frame-level features $X_f \in \mathbb{R}^{L \times d_f}$ and $M$ clip-level features $X_c \in \mathbb{R}^{M \times d_c}$, both from a pre-trained visual encoder.
Each branch is projected to a shared dimension~$d$, augmented with positional embeddings, and processed by our CurvSpec block:
\begin{equation}
\begin{aligned}
Z_f &= \mathrm{\textbf{CurvSpec}}\bigl(\mathrm{FC}(X_f) + P_f\bigr), \\
Z_c &= \mathrm{\textbf{CurvSpec}}\bigl(\mathrm{FC}(X_c) + P_c\bigr).
\end{aligned}
\end{equation}

\noindent\textbf{Similarity computation.}\quad
The retrieval score combines max-pooled cosine similarities at both granularities:
\begin{equation}
    S(q, V) = \alpha\;\max_{i}\;\cos(q, z_{f,i}) \;+\; (1{-}\alpha)\;\max_{j}\;\cos(q, z_{c,j}),
    \label{eq:score}
\end{equation}
where $\alpha \in [0,1]$ is a balancing hyperparameter.
This max-pooling strategy is particularly suited to the partial relevance setting, as it isolates the most semantically aligned moment in the video at each granularity rather than averaging over the full content, which would dilute the matching signal from the relevant segment.

\subsection{CurvSpec Block}
\label{sec:curvspec_block}

The CurvSpec block~(\cref{fig:methods}b) is our core component.
It processes input features through $N$ Euclidean attention layers and $N$ Lorentz attention layers, with each Lorentz layer using an independently learnable curvature, and fuses their outputs via a content-aware mechanism to capture flat relational structures and hierarchical semantics at multiple geometric scales.

\noindent\textbf{Euclidean attention.}\quad
Each Euclidean layer applies Gaussian-weighted self-attention~\cite{wang2023gmmformer}, with query, key, and value matrices obtained by learned projections from input $X \in \mathbb{R}^{L \times d}$.
A proximity mask $G_\sigma[i,j] = \mathcal{N}(j;\,i,\,\sigma^2)$ gates the temporal receptive field:
\begin{equation}
    H_\sigma = \mathrm{softmax}\!\left(G_\sigma \odot \frac{QK^\top}{\sqrt{d}}\right)V.
    \label{eq:eucl_attn}
\end{equation}
The first Euclidean layer uses global attention without Gaussian mask, the remaining $N{-}1$ layers use Gaussian widths $\sigma_n = 2^n$ for $n=1,\ldots,N{-}1$, yielding a multi-scale hierarchy of receptive fields.

\noindent\textbf{Lorentz attention.}\quad
Each Lorentz layer operates on $\mathbb{H}_{\kappa_n}^{d}$ with its own learnable curvature $\kappa_n$.
Input features are linearly projected, scaled by $\gamma \leq 1$ for numerical stability, and lifted onto $\mathbb{H}_{\kappa_n}^d$ via the exponential map.
Hyperbolic queries, keys, and values are computed via the Lorentz linear map, which maps to the tangent space, applies an affine transformation, and projects back:
\begin{equation}
    \mathrm{LL}(x) = \exp_o^\kappa\!\bigl(W\log_o^\kappa(x) + b\bigr).
    \label{eq:lorentz_linear}
\end{equation}
Denoting the resulting row vectors as $q_i$, $k_j$, $v_j^h$, the attention logits, weights, and Einstein midpoint aggregation~\cite{law2019lorentzian} yield:
\begin{equation}
\begin{aligned}
    a_{ij} &= \tfrac{\beta + \langle q_i,\, k_j\rangle_{\mathcal{L}}}{\tau},
    & w_{ij} &= \tfrac{e^{a_{ij}}}{\sum_{j'} e^{a_{ij'}}},\\[2pt]
    \bar{v}_i &= \textstyle\sum_{j=1}^{L} w_{ij}\,v_j^h,
    & [H_n]_i &= \tfrac{1}{\gamma}\log_o^{\kappa_n}\!\!\left(\tfrac{\bar{v}_i}{\sqrt{|\langle\bar{v}_i,\bar{v}_i\rangle_{\mathcal{L}}|}}\right),
\end{aligned}
    \label{eq:lorentz_block}
\end{equation}
where $[H_n]_i$ denotes the $i$-th row, $\tau$ a learnable temperature, and $\beta$ a constant bias centered at zero for coincident query--key pairs ($\kappa{=}1$).

All curvatures are initialized identically ($\kappa_n = 1.0$) and optimized independently.
During training, each layer converges to a distinct stable curvature spanning a wide geometric range: layers handling shallow structures settle near lower curvatures, while those capturing deep hierarchies gravitate toward higher values.
This emergent diversification, driven by the task objective, constitutes the learned \emph{curvature spectrum}.

\noindent\textbf{Content-aware fusion.}\quad
The $2N$ branch outputs $\{H_n\}_{n=1}^{2N}$ are combined via learned, input-dependent weights.
A global content token first queries each geometry branch to obtain branch-aware descriptors, which are then mapped to token-wise fusion weights:
\begin{equation}
\begin{aligned}
    g &= \frac{1}{L}\sum_{i=1}^{L}\frac{1}{2N}\sum_{n=1}^{2N} h_{i,n},
    & r_n &= \mathrm{Attn}(g, H_n, H_n),\\[2pt]
    \pi_{i,n} &= \mathrm{softmax}_{n}\!\left([W_2\phi(W_1 r_n)]_i/\tau_f\right),
    & o_i &= {\textstyle\sum}_{n=1}^{2N}\pi_{i,n} h_{i,n}.
\end{aligned}
    \label{eq:fusion}
\end{equation}
where $h_{i,n}$ is the $i$-th token from branch $H_n$, $g$ is the global content token, and $r_n$ is the branch descriptor obtained by attending to $H_n$.
The attention weights $\pi_{i,n}$ route each token toward its most suitable geometric representation.
Since different segments within a partially relevant video may exhibit fundamentally different semantic structures, this token-level selection allows the model to apply the most appropriate geometry to each position independently, rather than committing the entire sequence to a single fixed geometry.

\subsection{Semantic Centroid Decomposition}
\label{sec:centroids}

A central challenge in PRVR is that a query matches only a fraction of the video.
Representing a video as a single global vector conflates relevant and irrelevant segments.
We address this by decomposing each video into a set of \emph{semantic centroids}, each capturing a coherent thematic region, with the number of centroids adapted to the video's content complexity.

\noindent\textbf{Content-adaptive centroid count.}\quad
We quantify the semantic complexity of a video via the \emph{effective rank} of its clip feature matrix $X_c \in \mathbb{R}^{M \times d}$:
\begin{equation}
    r_{\mathrm{eff}}(X_c) = \exp\!\left(-\sum_j \bar{\sigma}_j \log \bar{\sigma}_j\right), \quad
    \bar{\sigma}_j = \frac{\sigma_j^2}{\sum_l \sigma_l^2},
    \label{eq:reff}
\end{equation}
where $\sigma_1 \geq \sigma_2 \geq \cdots$ are the singular values of $X_c$ in descending order.
$r_{\mathrm{eff}}$ measures the effective number of semantically independent directions in the clip feature space: small values indicate structurally simple videos dominated by a single theme, large values reflect multi-topic content with richer temporal diversity.
We select $K_i$ as the smallest value in $\{2, 4, 6\}$ for which the top-$K_i$ singular values collectively explain at least a fraction $\theta$ of the total variance:
\begin{equation}
K_i = \min \Bigg\{
K \in \{2, 4, 6\} \;\Bigg|\;
\frac{\sum_{j=1}^{K} \sigma_j^2}{\sum_{j} \sigma_j^2} \ge \theta
\Bigg\}.
\label{eq:adaptive_k}
\end{equation}
If no $K\in\{2,4,6\}$ satisfies the condition (rare in practice), $K_i$ defaults to $6$. The criterion is computed offline from pre-extracted features, incurring no inference overhead.
Crucially, $\theta$ is a single dataset-agnostic threshold: the same value applied to different corpora naturally yields different $K$ distributions, reflecting each corpus's intrinsic content diversity without per-dataset calibration (\cref{fig:adaptive_k}).

Given clip features $Z_c$, K-means clustering~\cite{macqueen1967kmeans} with $K_i$ clusters yields initial prototypes $\{\mu_k\}_{k=1}^{K_i}$.
Clip features are used for initialization because they provide coarser, semantically stable summaries at the scene level; frame-level details are then incorporated during the cross-attention refinement step.
Each prototype is then refined by cross-attending over both frame and clip features and fusing the results through a learned gate:
\begin{equation}
    c_k = \sigma(g_k) \odot \mathrm{Attn}(\mu_k, Z_f) \;+\; \bigl(1 {-} \sigma(g_k)\bigr) \odot \mathrm{Attn}(\mu_k, Z_c),
    \label{eq:centroid_refine}
\end{equation}
where $\sigma(g_k)$ is a sigmoid gate that adaptively modulates the contribution of each granularity:
\begin{equation}
    g_k \;=\; \mathrm{FC}\bigl([\,\mathrm{Attn}(\mu_k, Z_f) \;\|\; \mathrm{Attn}(\mu_k, Z_c)\,]\bigr),
    \label{eq:gate}
\end{equation}
with $[\cdot \| \cdot]$ denoting concatenation.
The refined centroids are projected onto $\mathbb{H}_\kappa^d$ via $\exp_o^\kappa$, and partial relevance is captured by the minimum geodesic distance to any centroid:
\begin{equation}
    d_\text{min}(q, V) = \min_{k \in [K_i]}\; d_\kappa\bigl(q_{\mathcal{L}},\, c_{k,\mathcal{L}}\bigr),
    \label{eq:dmin}
\end{equation}
where $q_{\mathcal{L}}$ and $c_{k,\mathcal{L}}$ denote the query and centroid projected onto the hyperboloid.
This allows the model to match a query against only the most relevant semantic region of a video, directly addressing the partial relevance challenge without requiring any temporal boundary annotations.

For videos with multiple queries, we further encourage diverse query--centroid correspondences by minimizing the average nearest-centroid distance:
\begin{equation}
    \mathcal{L}_\text{align} = \frac{1}{|\mathcal{V}|} \sum_{V \in \mathcal{V}} \frac{1}{N_V}\sum_{i=1}^{N_V} d_\text{min}(q_i, V),
    \label{eq:align}
\end{equation}
where $N_V$ is the number of queries associated with video $V$ and $\mathcal{V}$ denotes the subset of videos with at least two queries.
Combined with the diversity loss $\mathcal{L}_\text{div}$ (\cref{sec:optimization}), this encourages each centroid to specialize in a distinct semantic region.

\subsection{Training Objective}
\label{sec:optimization}

The overall objective unifies cross-modal matching in Euclidean space, centroid-based alignment in hyperbolic space, and geometric regularization:
\begin{equation}
    \mathcal{L} = \mathcal{L}_\text{match} + \lambda_h\,\mathcal{L}_\text{hyp} + \lambda_r\,\mathcal{L}_\text{reg}.
    \label{eq:total_loss}
\end{equation}

$\mathcal{L}_\text{match}$ aggregates bilateral NCE and triplet margin losses at both frame and clip granularities, encouraging the learned representations to rank positive query--video pairs above negatives.
$\mathcal{L}_\text{hyp}$ operates on the semantic centroids.
Its retrieval component follows an InfoNCE formulation over geodesic distances:
\begin{equation}
    \mathcal{L}_\text{retr} = -\frac{1}{B}\sum_{i=1}^{B}\log\frac{\exp\bigl({-}s \cdot d_\text{min}(q_i, V_i^+)\bigr)}{\sum_{j=1}^{B}\exp\bigl({-}s \cdot d_\text{min}(q_i, V_j)\bigr)},
    \label{eq:retr}
\end{equation}
where $V_i^+$ is the ground-truth video for query $q_i$ and $s$ is a learnable scaling factor.
The full term is $\mathcal{L}_\text{hyp} = \mathcal{L}_\text{retr} + \mathcal{L}_\text{align}$.
$\mathcal{L}_\text{reg}$ combines a diversity term $\mathcal{L}_\text{div}$ that prevents query embeddings from collapsing to nearby representations within a batch, and a curvature regularizer $\mathcal{L}_\text{curv}$ that confines all pairwise gaps $|\kappa_i - \kappa_j|$ within a target band $[\Delta_\text{min},\,\Delta_\text{max}]$, preventing both geometric collapse and uncontrolled divergence of the curvature spectrum.

\begin{table*}[t]
    \centering
    \small
    \renewcommand{\arraystretch}{0.84}
    \setlength{\tabcolsep}{4.8pt} 
    
    \begin{tabular}{l|ccccc|ccccc|ccccc}
    \toprule
    \textbf{Method} &
    \multicolumn{5}{c|}{\textbf{ActivityNet Captions}} &
    \multicolumn{5}{c|}{\textbf{TVR}} &
    \multicolumn{5}{c}{\textbf{Charades-STA}} \\
    \cmidrule(lr){2-6} \cmidrule(lr){7-11} \cmidrule(lr){12-16}
     & R@1 & R@5 & R@10 & R@100 & SumR
     & R@1 & R@5 & R@10 & R@100 & SumR
     & R@1 & R@5 & R@10 & R@100 & SumR \\
    \midrule
    
    \multicolumn{16}{l}{\textit{Text-to-Video Retrieval}} \\
    HGR~\cite{Chen2020FineGrainedVR} & 4.0 & 15.0 & 24.8 & 63.2 & 107.0 & 1.7 & 4.9 & 8.3 & 35.2 & 50.1 & 1.2 & 3.8 & 7.3 & 33.4 & 45.7 \\
    DE++~\cite{dong2022dual_encoding} & 5.3 & 18.4 & 29.2 & 68.0 & 121.0 & 8.8 & 21.9 & 30.2 & 67.4 & 128.3 & 1.7 & 5.6 & 9.6 & 37.1 & 54.1 \\
    RIVRL~\cite{RIVRL} & 5.2 & 18.0 & 28.2 & 66.4 & 117.8 & 9.4 & 23.4 & 32.2 & 70.6 & 135.6 & 1.6 & 5.6 & 9.4 & 37.7 & 54.3 \\
    CLIP4Clip~\cite{luo2022clip4clip} & 5.9 & 19.3 & 30.4 & 71.6 & 127.3 & 9.9 & 24.3 & 34.3 & 72.5 & 141.0 & 1.8 & 6.5 & 10.9 & 44.2 & 63.4 \\
    Cap4Video~\cite{wu2023cap4video} & 6.3 & 20.4 & 30.9 & 72.6 & 130.2 & 10.3 & 26.4 & 36.8 & 74.0 & 147.5 & 1.9 & 6.7 & 11.3 & 45.0 & 65.0 \\
    \addlinespace
    \hline
    \addlinespace
    
    \multicolumn{16}{l}{\textit{Video Corpus Moment Retrieval (w/o moment localization)}} \\
    XML~\cite{lei2020tvr} & 5.3 & 19.4 & 30.6 & 73.1 & 128.4 & 10.7 & 28.1 & 38.1 & 80.3 & 157.1 & 1.6 & 6.0 & 10.1 & 46.9 & 64.6 \\
    ReLoCLNet~\cite{zhang2021reloclnet} & 5.7 & 18.9 & 30.0 & 72.0 & 126.6 & 10.0 & 26.5 & 37.3 & 81.3 & 155.1 & 1.2 & 5.4 & 10.0 & 45.6 & 62.3 \\
    CONQUER~\cite{hou2021conquer} & 6.5 & 20.4 & 31.8 & 74.3 & 133.1 & 11.0 & 28.9 & 39.6 & 81.3 & 160.8 & 1.8 & 6.3 & 10.3 & 47.5 & 66.0 \\
    JSG~\cite{chen2023jsg} & 6.8 & 22.7 & 34.8 & 76.1 & 140.5 & -- & -- & -- & -- & -- & 2.4 & 7.7 & 12.8 & 49.8 & 72.7 \\
    \addlinespace
    \hline
    \addlinespace
    
    \multicolumn{16}{l}{\textit{Partially Relevant Video Retrieval}} \\
    MS-SL~\cite{dong2022prvr} & 7.1 & 22.5 & 34.7 & 75.8 & 140.1 & 13.5 & 32.1 & 43.4 & 83.4 & 172.4 & 1.8 & 7.1 & 11.8 & 47.7 & 68.4 \\
    PEAN~\cite{Jiang2023ProgressiveEA} & 7.4 & 23.0 & 35.5 & 75.9 & 141.8 & 13.5 & 32.8 & 44.1 & 83.9 & 174.2 & \textbf{2.7} & 8.1 & 13.5 & 50.3 & 74.7 \\
    LH~\cite{LH} & 7.4 & 23.5 & 35.8 & 75.8 & 142.4 & 13.2 & 33.2 & 44.4 & 85.5 & 176.3 & 2.1 & 7.5 & 12.9 & 50.1 & 72.7 \\
    GMMFormer~\cite{wang2023gmmformer} & 8.3 & 24.9 & 36.7 & 76.1 & 146.0 & 13.9 & 33.3 & 44.5 & 84.9 & 176.6 & 2.1 & 7.8 & 12.5 & 50.6 & 72.9 \\
    ARL~\cite{cho2025arl} & 8.3 & 24.6 & 37.4 & 78.0 & 148.3 & 15.6 & 36.3 & 47.7 & 86.3 & 185.9 & -- & -- & -- & -- & -- \\
    MamFusion~\cite{ying2025mamfusion} & 8.0 & 25.4 & 37.2 & 76.8 & 147.4 & 14.2 & 33.9 & 44.9 & 84.5 & 177.5 & -- & -- & -- & -- & -- \\
    ProtoPRVR~\cite{moon2025protoprvr} & 7.9 & 24.9 & 37.2 & 77.4 & 147.4 & 15.4 & 35.9 & 47.5 & 86.3 & 185.1 & -- & -- & -- & -- & -- \\
    HLFormer~\cite{Li25_HLFormer} & \underline{8.7} & \underline{27.1} & \textbf{40.1} & \underline{79.0} & \underline{154.9} & \underline{15.7} & \underline{37.1} & \underline{48.5} & \underline{86.4} & \underline{187.7} & \underline{2.6} & \underline{8.5} & \underline{13.7} & \textbf{54.0} & \underline{78.7} \\
    
    \addlinespace
    \hline
    \addlinespace
    
    \rowcolor{gray!12}
    \textbf{CurvSpec (Ours)} & \textbf{9.0} & \textbf{27.6} & \underline{39.9} & \textbf{79.1} & \textbf{155.6} & \textbf{15.9} & \textbf{37.5} & \textbf{49.0} & \textbf{86.7} & \textbf{189.1} & \underline{2.6} & \textbf{8.9} & \textbf{14.3} & \underline{53.3} & \textbf{79.1} \\
    \bottomrule
    \end{tabular}
    \vspace{3mm}
    \caption{\textbf{Retrieval performance on ActivityNet Captions, TVR, and Charades-STA.} \textbf{Bold} and \underline{underline} denote the best and second-best results, respectively. ``--'' indicates unavailable results.}
    \label{tab:three_benchmarks}
    \end{table*}
    

    \section{Experiments}
    \subsection{Experimental Setup}
    
    \vspace{1mm}\noindent \textbf{Datasets.}
    We evaluate our model on three widely used large-scale video datasets that span diverse content domains: ActivityNet Captions~\cite{krishna2017densecaptioningeventsvideos}, TVR~\cite{lei2020tvr}, and Charades-STA~\cite{gao2017talltemporalactivitylocalization}.
    \textbf{ActivityNet Captions} was originally introduced for dense video captioning and has since become a standard benchmark for partially relevant video retrieval. It consists of approximately 20K YouTube videos with an average duration of around 118 seconds.
    \textbf{TVR} comprises around 21.8K video clips collected from six different TV shows, with an average duration of roughly 76 seconds per video. Each clip is associated with five natural language sentences that describe distinct moments within the video.
    \textbf{Charades-STA} includes 6,670 videos and 16,128 sentence-level annotations.
    For all datasets, we follow the standard training and testing splits adopted in previous works.

    \vspace{1mm}\noindent \textbf{Evaluation Strategy.} Following prior work~\cite{wang2023gmmformer, Li25_HLFormer, zhang2025enhancingpartiallyrelevantvideo}, we adopt rank-based evaluation metrics, including Recall at K (R@K, K = 1,5,10,100).
    R@K measures the proportion of test queries whose corresponding items appear within the top K retrieved results.
    For comprehensive comparison, we also calculate the Sum of Recalls (SumR), which aggregates all R@K scores.
    
    \vspace{1mm}\noindent \textbf{Implementation Details.}~Following standard practice, we use pre-extracted visual features: ResNet-152+I3D (3072-dim) for TVR~\cite{lei2020tvr}, and I3D for ActivityNet and Charades~\cite{zhang2021reloclnet,mun2020localglobalvideotextinteractionstemporal}. Text features are RoBERTa embeddings~\cite{liu2019robertarobustlyoptimizedbert}: 768-dim for TVR, 1024-dim for ActivityNet and Charades~\cite{dong2022prvr}.

    Each CurvSpec block contains $2N{=}8$ parallel attention layers: $N{=}4$ Euclidean layers and $N{=}4$ Lorentz layers, with each Lorentz layer using an independently learnable curvature.
    Following the design strategy of GMMFormer~\cite{wang2023gmmformer}, we employ multiple Gaussian widths to capture contextual dependencies at different scales, using one global layer and widths ranging from \(2^{1}\) to \(2^{N-1}\).

\begin{figure}[t]
        \centering
        \includegraphics[width=\linewidth]{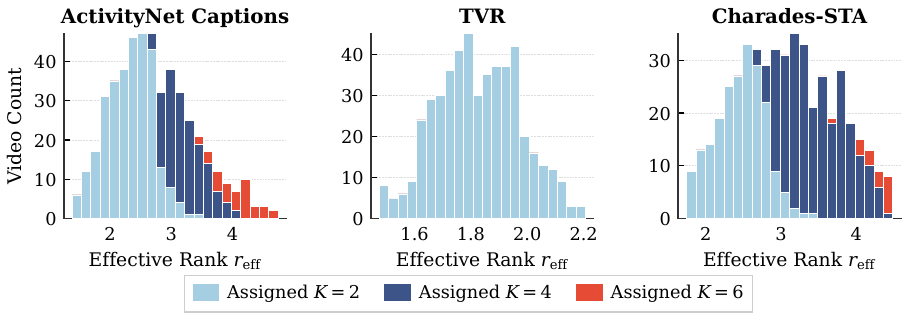}
        \caption{\textbf{Adaptive centroid count $K_i$ across datasets}. Stacked histograms of per-video effective rank $r_{\mathrm{eff}}$ (\cref{eq:reff}), coloured by the assigned $K_i \in \{2, 4, 6\}$. }
        \label{fig:adaptive_k}
    \end{figure}

    For semantic centroid initialization, we employ K-means with the content-adaptive $K_i$ defined in~\cref{sec:centroids}. The variance-explanation threshold is set to $\theta{=}0.85$, selected on the Charades-STA validation set. As shown in~\cref{fig:adaptive_k}, this single threshold produces naturally different $K$ distributions across datasets, with structurally simpler corpora (e.g., TVR) predominantly assigned $K{=}2$ and more diverse ones (ActivityNet, Charades-STA) receiving higher values, confirming that the criterion captures intrinsic content diversity without per-dataset tuning.

    For the training configurations, we train the model using the Adam optimizer with a batch size of 128.
    All experiments are implemented in PyTorch and conducted on NVIDIA RTX 3090 GPU.

    \begin{table}[t]
    \centering
    \small
    \renewcommand{\arraystretch}{0.85}
    \setlength{\tabcolsep}{3pt}
    \begin{tabular}{l|ccc|cc}
    \toprule
    \textbf{Method} & \textbf{Params} & \textbf{FLOPs} & \textbf{Latency}$^{\dagger}$ & \textbf{R@1} & \textbf{SumR} \\
    \midrule
    GMMFormer & 11.0M & 1.63G & 2.58ms & 13.9 & 176.6 \\
    HLFormer  & 28.0M & 4.73G & 2.90ms & 15.7 & 187.7 \\
    \rowcolor{gray!14}
    \textbf{CurvSpec} & 28.8M & 4.74G & 2.96ms & 15.9 & 189.1 \\
    \bottomrule
    \end{tabular}
    \vspace{3mm}
    \caption{\textbf{Efficiency-accuracy tradeoff} on TVR (RTX 3090). $^{\dagger}$Online query latency (ms/query).}
    \label{tab:efficiency}
    \end{table}

    \begin{table*}[t]
    \centering
    \small
    \renewcommand{\arraystretch}{0.8}
    \setlength{\tabcolsep}{3.8pt}
    
    \begin{tabular}{l|ccccc|ccccc|ccccc}
    \toprule
    \textbf{Configuration} &
    \multicolumn{5}{c|}{\textbf{ActivityNet Captions}} &
    \multicolumn{5}{c|}{\textbf{TVR}} &
    \multicolumn{5}{c}{\textbf{Charades-STA}} \\
    \cmidrule(lr){2-6} \cmidrule(lr){7-11} \cmidrule(lr){12-16}
    & R@1 & R@5 & R@10 & R@100 & SumR
    & R@1 & R@5 & R@10 & R@100 & SumR
    & R@1 & R@5 & R@10 & R@100 & SumR \\
    \midrule
    
    \multicolumn{16}{l}{\textit{Geometric Modeling Strategy}} \\
    \quad Euclidean Only & 8.3 & 26.2 & 38.5 & 78.5 & 151.5 & 14.9 & 36.4 & 47.7 & 86.2 & 185.2 & 2.4 & 8.1 & 13.5 & 52.1 & 76.1 \\
    \quad Fixed Curvature & 8.4 & 26.9 & 39.6 & 78.6 & 153.5 & 15.4 & 37.1 & 48.4 & 86.2 & 186.9 & 2.4 & 8.3 & 13.7 & 52.8 & 77.2 \\
    \addlinespace
    \hline
    \addlinespace
    
    \multicolumn{16}{l}{\textit{Aggregation Strategy}} \\
    \quad Tangent Space Sum & 8.6 & 26.9 & 39.2 & 78.1 & 152.8 & 15.3 & 37.0 & 48.5 & 86.0 & 186.8 & 2.3 & 8.4 & 13.7 & 52.9 & 77.3 \\
    \addlinespace
    \hline
    \addlinespace
    
    \multicolumn{16}{l}{\textit{Partial Relevance Modeling}} \\
    \quad w/o Semantic Centroids & 8.3 & 26.0 & 39.1 & 78.5 & 151.9 & 15.7 & 35.8 & 47.0 & 85.6 & 184.1 & 2.3 & 8.5 & 13.4 & 51.3 & 75.5 \\
    \addlinespace
    \hline
    \addlinespace
    
    \multicolumn{16}{l}{\textit{Feature Granularity}} \\
    \quad w/o Frame Features & 8.0 & 24.8 & 37.1 & 76.9 & 146.8 & 14.1 & 34.0 & 44.9 & 85.1 & 178.1 & 2.1 & 8.3 & 14.0 & 52.1 & 76.5 \\
    \quad w/o Clip Features & 7.6 & 23.0 & 35.3 & 75.4 & 141.3 & 12.6 & 31.8 & 43.3 & 83.4 & 171.1 & 1.9 & 8.0 & 13.5 & 50.1 & 73.5 \\
    \addlinespace
    \hline
    \addlinespace
    
    \multicolumn{16}{l}{\textit{Auxiliary Losses}} \\
    \quad w/o $\mathcal{L}_{\text{reg}}$ & 8.7 & 27.4 & 39.5 & 78.3 & 153.9 & 15.4 & 35.8 & 47.6 & 86.5 & 185.3 & 2.4 & 8.5 & 14.0 & 52.5 & 77.4 \\
    \quad w/o $\mathcal{L}_{\text{div}}$ & 8.6 & 27.0 & 39.4 & 77.9 & 152.9 & 15.4 & 36.0 & 47.9 & 86.1 & 185.4 & 2.2 & 8.5 & 14.0 & 52.2 & 76.9 \\
    \addlinespace
    \hline
    \addlinespace
    \rowcolor{gray!12}
    \textbf{CurvSpec (Full)} & \textbf{9.0} & \textbf{27.6} & \textbf{39.9} & \textbf{79.1} & \textbf{155.6} & \textbf{15.9} & \textbf{37.5} & \textbf{49.0} & \textbf{86.7} & \textbf{189.1} & \textbf{2.6} & \textbf{8.9} & \textbf{14.3} & \textbf{53.3} & \textbf{79.1} \\
    \bottomrule
    \end{tabular}
    \vspace{3mm}
    \caption{\textbf{Ablation study} on ActivityNet Captions, TVR, and Charades-STA. Each group isolates one design dimension.}
    \label{tab:ablation_study}
    \end{table*}

    \subsection{Comparison with Prior Works}
    
    \vspace{1mm}\noindent \textbf{Baselines.}
    For the PRVR task, we compare with eight representative methods: MS-SL~\cite{dong2022prvr}, PEAN~\cite{Jiang2023ProgressiveEA}, LH~\cite{LH}, GMMFormer~\cite{wang2023gmmformer}, ARL~\cite{cho2025arl}, MamFusion~\cite{ying2025mamfusion}, ProtoPRVR~\cite{moon2025protoprvr}, and HLFormer~\cite{Li25_HLFormer}.
    We further compare with models from related tasks:
    for T2VR, HGR~\cite{Chen2020FineGrainedVR}, DE++~\cite{dong2022dual_encoding}, RIVRL~\cite{RIVRL}, CLIP4Clip~\cite{luo2022clip4clip}, and Cap4Video~\cite{wu2023cap4video};
    for VCMR, XML~\cite{lei2020tvr}, ReLoCLNet~\cite{zhang2021reloclnet}, CONQUER~\cite{hou2021conquer}, and JSG~\cite{chen2023jsg}.

    \vspace{1mm}\noindent \textbf{Performance Comparison.} 
    As shown in~\cref{tab:three_benchmarks}, CurvSpec achieves the best performance on the majority of metrics across all three benchmarks, with especially consistent gains in aggregate SumR. CurvSpec also outperforms recent PRVR methods such as MamFusion and ProtoPRVR under the same I3D-based protocol. In particular, CurvSpec ranks first in SumR on all datasets, demonstrating the combined effectiveness of our learnable curvature spectrum and semantic centroid decomposition for partial relevance modeling.

    \vspace{1mm}\noindent \textbf{Efficiency Analysis.}
    \cref{tab:efficiency} reports the efficiency-accuracy tradeoff on TVR. Despite introducing mixed-geometry attention and semantic centroid decomposition, CurvSpec maintains comparable FLOPs and online query latency to existing methods, while achieving the best retrieval accuracy. The adaptive-$K$ selection and K-means initialization are performed offline, so they do not affect online retrieval latency. This confirms the geometric operations in CurvSpec add negligible computational overhead, making the approach deployable in real-time retrieval scenarios.

    \begin{figure}[tbp]
        \centering
        \begin{subfigure}[b]{0.50\linewidth}
            \includegraphics[width=\linewidth]{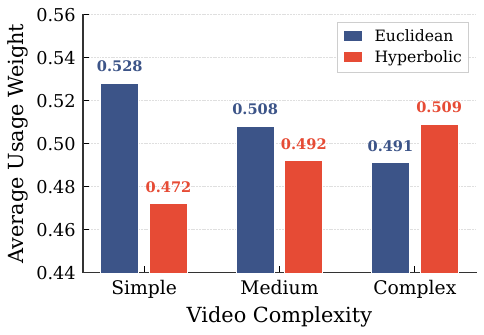}
            \caption{Geometry usage distribution}
            \label{fig:geometry_usage}
        \end{subfigure}
        \hspace{0.01\linewidth}
        \begin{subfigure}[b]{0.46\linewidth}
            \includegraphics[width=\linewidth]{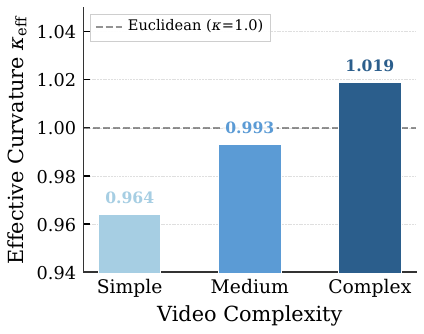}
            \caption{Curvature distribution}
            \label{fig:effective_curvature}
        \end{subfigure}
        \caption{\textbf{Adaptive geometric selection.} (a) Videos with higher complexity utilize more hyperbolic geometry. (b) The effective curvature $\kappa_{\text{eff}}$ increases with video complexity.}
        \label{fig:adaptive_geometry}
    \end{figure}

    \subsection{Ablation Study}
    \label{sec:ablation_study}
    We conduct systematic ablation studies on all three benchmarks to isolate and evaluate the contribution of each individual component. Results are presented in~\cref{tab:ablation_study}.

    \vspace{1mm}\noindent \textbf{Geometric Modeling Strategy.}
    We compare three variants: (i) replacing all Lorentz attention layers with standard Euclidean attention, (ii) using a single fixed curvature across all hyperbolic layers, and (iii) our learnable curvature spectrum. The learnable spectrum consistently outperforms both alternatives, confirming that content-adaptive geometric biases provide richer representational capacity than any single fixed geometry.

    \begin{figure}[tbp]
        \centering
        \includegraphics[width=\linewidth,height=0.43\linewidth]{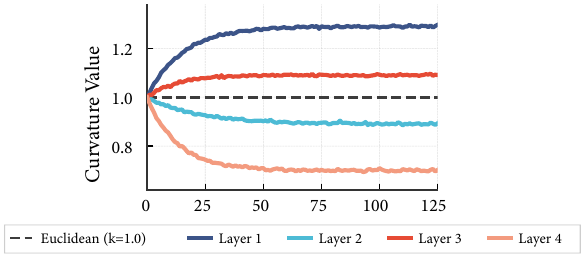}
        \caption{Evolution of \textbf{learned curvatures} during training. Different attention layers converge to distinct stable values, demonstrating the emergence of a geometric spectrum.}
        \label{fig:curvature_evolution}
    \end{figure}

    \begin{figure*}[tbp]
        \centering
        \includegraphics[width=0.85\linewidth]{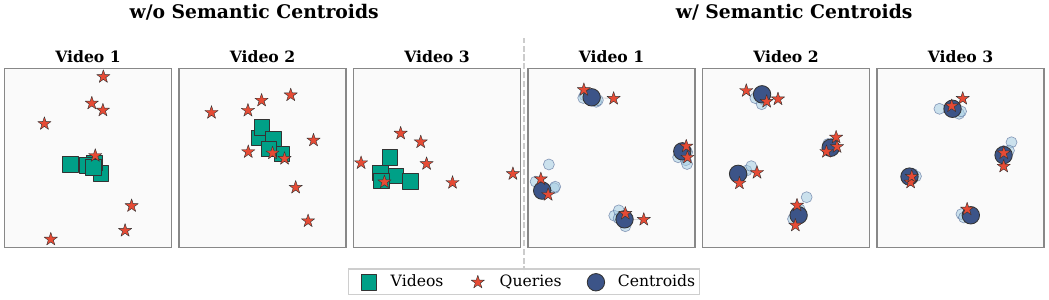}
        \caption{\textbf{t-SNE visualization of query and video embeddings} for three video examples from TVR.}
        \label{fig:centroid_visualization}
    \end{figure*}

    \vspace{1mm}\noindent \textbf{Aggregation Strategy.}
    In our Lorentz attention, values are aggregated via the Einstein midpoint directly on the hyperboloid. We compare this against an alternative that maps values to the tangent space, performs weighted summation, and projects back. The Einstein midpoint yields consistent improvements across all datasets, as tangent space operations introduce geometric distortions that accumulate in high-curvature regions, progressively undermining the fidelity of the resulting hyperbolic representations.
    
    \vspace{1mm}\noindent \textbf{Partial Relevance Modeling.}
    Removing semantic centroids and reverting to globally pooled video representations leads to consistent and notable performance drops across all three datasets. This validates that decomposing videos into multiple semantic regions enables more precise query-to-segment matching.

    \vspace{1mm}\noindent \textbf{Feature Granularity.}
    Both frame-level and clip-level features contribute to the full model, with clip features proving more critical overall, as removing them causes noticeably larger degradation. This supports our coarse-to-fine design where clip features capture global semantic context and frame features encode fine-grained temporal dynamics.
    
    \vspace{1mm}\noindent \textbf{Auxiliary Losses.}
    Both $\mathcal{L}_{\text{reg}}$ and $\mathcal{L}_{\text{div}}$ contribute to the final performance. While their individual metric gains appear modest, $\mathcal{L}_{\text{reg}}$ plays a crucial role in preventing curvature collapse during training, and $\mathcal{L}_{\text{div}}$ encourages discriminative query embeddings.

    \begin{table}[tbp]
    \centering
    \small
    \renewcommand{\arraystretch}{0.85}
    \setlength{\tabcolsep}{2.7pt}
    \begin{tabular}{l|cc|cc|cc}
    \toprule
    & \multicolumn{2}{c|}{\textbf{Simple}} & \multicolumn{2}{c|}{\textbf{Medium}} & \multicolumn{2}{c}{\textbf{Complex}} \\
    \cmidrule(lr){2-3} \cmidrule(lr){4-5} \cmidrule(lr){6-7}
    \textbf{Variant} & R@1 & SumR & R@1 & SumR & R@1 & SumR \\
    \midrule
    Euclidean Only & 17.2 & 191.5 & 15.1 & 186.0 & 11.8 & 175.8 \\
    Fixed Curvature & 17.4 & 192.0 & 15.6 & 187.8 & 12.5 & 178.4 \\
    \rowcolor{gray!14}
    \textbf{CurvSpec} & \textbf{17.6} & \textbf{192.8} & \textbf{16.0} & \textbf{189.6} & \textbf{13.6} & \textbf{183.2} \\
    \midrule
    $\Delta$ {\scriptsize(vs.\ Euclidean)} & {+0.4} & {+1.3} & {+0.9} & {+3.6} & \textbf{+1.8} & \textbf{+7.4} \\
    \bottomrule
    \end{tabular}
    \vspace{3mm}
    \caption{\textbf{Per-complexity retrieval performance} on TVR. }
    \label{tab:per_complexity}
    \end{table}

    \subsection{Analysis}
    \label{sec:analysis}
    
    \vspace{1mm}\noindent \textbf{Learned Curvature Diversity.}
    \cref{fig:curvature_evolution} tracks the curvature parameters of each Lorentz attention layer across training epochs. Starting from identical initialization, all layers converge to distinct and stable values that span a wide geometric range, showing that the model discovers a heterogeneous curvature spectrum instead of collapsing to a single geometry. This diversification emerges from the retrieval objective without explicit geometric supervision, suggesting that different types of semantic relationships inherently favor different curvature regimes.
    
    \vspace{1mm}\noindent \textbf{Content-Aware Geometric Adaptation.}
    We examine how the learned spectrum is applied to videos of varying structural complexity. As shown in~\cref{fig:adaptive_geometry}(a), the adaptive fusion mechanism assigns progressively greater weight to hyperbolic layers as video complexity increases, while simpler videos are predominantly processed in near-Euclidean geometry. Correspondingly, \cref{fig:adaptive_geometry}(b) reveals that the effective curvature $\kappa_{\text{eff}}$ exhibits a clear positive correlation with complexity, consistent with the principle of content-aware geometric allocation. This behavior aligns with the theoretical motivation that hyperbolic spaces, through exponential volume growth, are better suited for encoding deep hierarchies, while flat Euclidean geometry suffices for shallow semantic structures.
    
    \vspace{1mm}\noindent \textbf{Per-Complexity Performance Breakdown.}
    To validate that this adaptive behavior translates to tangible retrieval gains, we stratify the TVR test set by video complexity and compare geometric modeling variants in~\cref{tab:per_complexity}.
    We define a video complexity score $\mathcal{C}$ by combining three clip-level statistics: temporal feature variance $\sigma_t^2$, pairwise cosine distance diversity $\sigma_d$, and effective rank $r_{\text{eff}}$ (\cref{eq:reff}).
    Videos are stratified into Simple ($\mathcal{C} \leq \text{P25}$), Medium, and Complex ($\mathcal{C} > \text{P75}$) groups.
    While CurvSpec yields consistent improvements across all groups, the gains are disproportionately concentrated on structurally complex videos, where the SumR improvement is over $5{\times}$ larger than on simple ones. This provides direct evidence that learnable curvature allocates stronger geometric biases precisely where hierarchical depth demands richer modeling capacity.
    
    \vspace{1mm}\noindent \textbf{Semantic Centroid Visualization.}
    \cref{fig:centroid_visualization} visualizes the embedding space via t-SNE. With semantic centroids (right panel), query embeddings cluster tightly around their matched video regions, forming well-separated and compact groups that enable fine-grained partial relevance matching. In contrast, the global-pooling baseline (left panel) produces scattered query embeddings far from the coarse video representations, reflecting the inherent difficulty of aligning a specific text description against an entire untrimmed video. \cref{fig:adaptive_k} further confirms that the adaptive criterion assigns richer decompositions to structurally complex videos and fewer centroids to uniform ones, achieving fine-grained matching with negligible additional overhead.

\section{Conclusion}

We presented CurvSpec, a framework for partially relevant video retrieval that combines a learnable curvature spectrum with semantic centroid decomposition. The curvature spectrum processes features across multiple Euclidean and hyperbolic attention layers whose curvatures are optimized end-to-end, while the semantic centroids decompose each video into coherent regions for fine-grained matching. Experiments on three benchmarks show state-of-the-art performance with comparable efficiency to single-geometry baselines. 
Our work highlights a broader insight for video retrieval that partial relevance should not be treated only as a problem of finding the right temporal evidence, it also depends on whether the retrieval space has the right geometry for the video's semantic structure.
Learning curvature from content offers a principled way to align representation geometry with the heterogeneous structures of real-world videos.

\begin{acks}
This study was conducted entirely during Zhen Liu's internship at Tsinghua University and was supported by the National Natural Science Foundation of China (Grant Nos. 92467204 and 62472249), the Shenzhen Science and Technology Program (Grant Nos. KJZD\allowbreak20240903102300001 and JCYJ20250604145014018), the Natural Science Foundation for Top Talents of Shenzhen Technology University (Grant No. GDRC202413), and the National Natural Science Foundation of China under Grant 624B2088.
\end{acks}

\bibliographystyle{ACM-Reference-Format}
\bibliography{ref}

\end{document}